\documentclass[a4paper,conference]{IEEEtran}
\ifCLASSINFOpdf
\else
\fi
\usepackage[pdftex]{graphicx}
\usepackage{cite}
\usepackage{amsmath,amssymb,amsfonts}
\usepackage{algorithmic}
\usepackage{textcomp}
\usepackage{xcolor}
\usepackage{pifont}
\usepackage{multirow}

\usepackage{hyperref}
\usepackage{booktabs}

\begin{document}

%
\title{Leveraging Synthetic Data in Object Detection
on Unmanned Aerial Vehicles}

\author{\IEEEauthorblockN{Benjamin Kiefer\IEEEauthorrefmark{1}}
\IEEEauthorblockA{University of Tuebingen\\
benjamin.kiefer@uni-tuebingen.de}
\and
\IEEEauthorblockN{David Ott\IEEEauthorrefmark{1}}
\IEEEauthorblockA{University of Tuebingen\\
david.ott@uni-tuebingen.de}
\and
\IEEEauthorblockN{Andreas Zell}
\IEEEauthorblockA{University of Tuebingen\\
andreas.zell@uni-tuebingen.de}}


%


\maketitle
\begingroup\renewcommand\thefootnote{*}
\footnotetext{Equal contribution. This work has been supported by the German Ministry for Economic Affairs and Energy, Project Avalon, FKZ: 03SX481B.}

\endgroup
\begin{abstract}
Acquiring data to train deep learning-based object detectors on Unmanned Aerial Vehicles (UAVs) is expensive, time-consuming and may even be prohibited by law in specific environments. On the other hand, synthetic data is fast and cheap to access.
In this work, we explore the potential use of synthetic data in object detection from UAVs across various application environments. For that, we extend the open-source framework DeepGTAV to work for UAV scenarios. We capture various large-scale high-resolution synthetic data sets in several domains to demonstrate their use in real-world object detection from UAVs by analyzing multiple training strategies across several models. Furthermore, we analyze several different data generation and sampling parameters to provide actionable engineering advice for further scientific research. The DeepGTAV framework is available at \url{https://git.io/Jyf5j}.
                        
\end{abstract}


%
\IEEEpeerreviewmaketitle

\section{Introduction}

Unmanned Aerial Vehicles (UAVs) equipped with cameras are increasingly used as autonomous vision systems in a wide range of applications, such as traffic surveillance, search and rescue, agriculture and smart cities \cite{adao2017hyperspectral,lygouras2019unsupervised,geraldes2019uav,san2018uav}. In all of these scenarios, these systems rely on the robust detection of objects of interest. Although object detection on natural images taken from hand-held or car-mounted cameras has been studied intensively, object detection from UAVs trails behind in performance \cite{zhu2018visdrone}. This is partly due to the limited amount of annotated publicly available data sets. In turn, this is partly caused by the highly complex data generating missions, which are subject to permissions, UAV flying restrictions and environmental factors \cite{varga2021seadronessee}. Furthermore, there are more degrees of freedom in the UAV domain (camera angles, position), which account for objects from unnatural perspectives, e.g. small objects from above. These difficulties are on top of other common obstacles, such as high and enduring labeling costs.

With more publicly available data sets, object detection on UAVs could be improved. However, data collection and labeling are expensive and time-consuming. Furthermore, data set collection raises serious privacy (e.g. GDPR \cite{GDPR} in Europe) \cite{msceleb} and security concerns because specific locations demand a long and complicated approval procedure.

Moreover, currently published data sets suffer from large class and domain imbalances \cite{du2018unmanned,zhu2018visdrone,kiefer2021leveraging,messmer2021gaining}. Both problems are inherently caused by a problematic capturing procedure, where many variables cannot be controlled. 

On the other hand, synthetically generated data for computer vision problems can help train data demanding visual perception systems because it is comparably fast and inexpensive to acquire. Furthermore, this data can easily be tailored to specific requirements. Several works address synthetic data generation in computer vision. However, most of them focus on driving, simulating constrained traffic situations \cite{kar2019meta,srivastava2019multi,hurl2019precise}. Few works consider the generation of synthetic data captured from UAVs. However, these only focus on the capturing process of the sensors and are lacking in world, object and physics details, or do not feature them at all \cite{kar2019meta,Fonder2019MidAir}.

\begin{figure}
	\centering
	\begin{tabular}{@{}c@{\hspace{.49cm}}c@{}}
		\includegraphics[trim=0 0 0 0,clip,width=.49\textwidth]{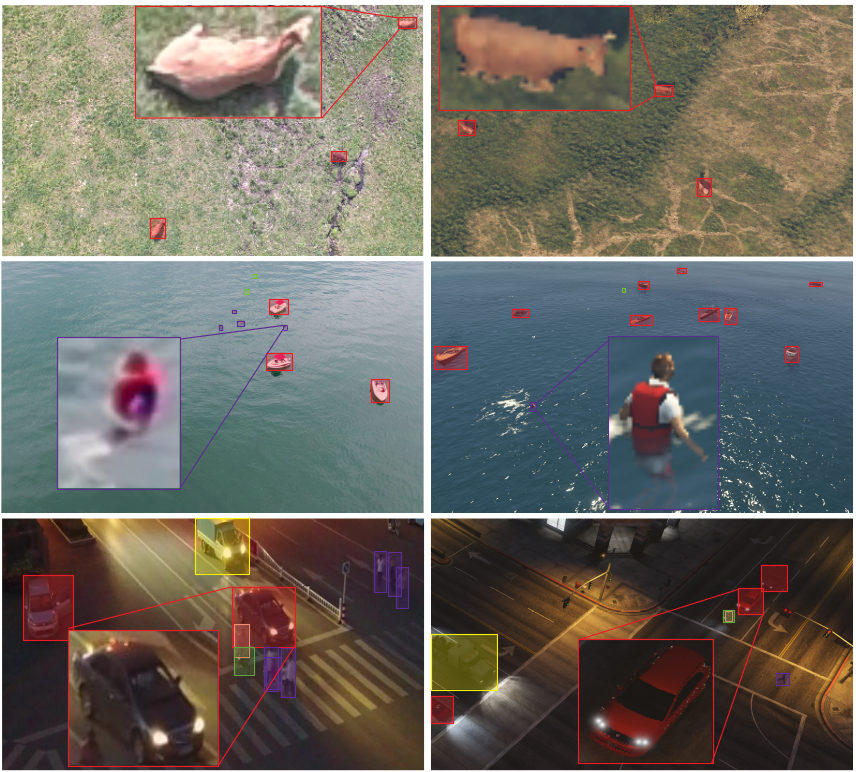}
		
		
	\end{tabular}
	\caption{Real (left) and synthetic (right) image samples in different applications scenarios with ground truth annotations. Representative objects are magnified.}
	\label{fig:intro_img}
\end{figure}


While the use of synthetic data in the context of autonomous driving has been investigated thoroughly, it is not clear whether these findings can be applied in the UAV setting. As mentioned in the beginning, the problems for general UAV object detection also apply to synthetic data generation. Most simulation engines focus on autonomous driving, and therefore rendering is aimed at looking realistic for these scenarios. Even if these simulation engines could technically be adapted to the UAV setting, light conditions, shadows, visibility range, rendered resolution, and more may significantly affect the quality of the rendered footage. In turn, the generated data may be less valuable for transfer to the real world. 

In this work, we consider the video game Grand Theft Auto V (GTAV) \cite{gtavcite} as a simulation platform. It offers numerous detailed object models that interact in a large world with realistic graphics and physics simulations. Building on previous works \cite{johnson2016driving,angus2018unlimited,yue2018lidar,hurl2019precise}, we extend the DeepGTAV data extraction tool to work for airborne scenarios by extending its functionalities regarding in-game agent and camera position and rotation, environment manipulation, object spawning and metadata extraction.

Using this simulation engine, we create three large-scale high-resolution (4K) synthetic object detection data sets in different application scenarios. Using these, we evaluate different training strategies and their transfer performances on three corresponding real-world data sets. We provide analytical insights and actionable advice on which settings to choose by doing extensive experiments and ablations.



In particular, our contributions are as follows:

\begin{itemize}
	\item We modify and extend the DeepGTAV tool, in particular, to allow the production of airborne data. We discuss the specific improvements in Section \ref{sec:ToolDescriptionAndImprovements}.
	
	\item We provide three large-scale high-resolution metadata annotated data sets in different UAV application scenarios and make them publicly available.
	
	\item We evaluate the applicability of these data sets to improve real-world object detection and analyze the from-scratch performance.
	
	\item We analyze the influence of different parameters of the data generation, e.g. the graphics quality and the alignment of metadata.

\end{itemize}

\subsection{Related Work}

This section reviews the literature regarding synthetic data generation engines, synthetic data taken from UAVs and the role of meta/environmental data in this context.

\subsubsection{Synthetic Data Generation Engines}
Many data generation engines focus on autonomous driving. AirSim \cite{shah2018airsim} and Carla \cite{dosovitskiy2017carla} leverage the Unreal Engine \cite{qiu2016unrealcv} to create a simulation engine suitable for autonomous driving scenarios. While AirSim also provides support for UAV scenarios, it lacks a world to simulate objects and its physics implementation, which must be created and modeled first.
Instead of laboriously creating simulation engines, some researchers use the computer game GTAV \cite{gtavcite} to generate data. With DeepGTAV \cite{johnson2016driving}, researchers leverage GTAV to gather data for autonomous driving. PreSIL \cite{hurl2019precise} builds upon this approach to refine the data acquisition process with their tool DeepGTAV-PreSIL. However, both systems lack the feature to modify the in-game agent and camera position and rotation, essentially limiting it to a single autonomous driving scenario. Further manipulations, such as spawning objects, manipulating the environment and extracting the metadata, are not possible either. The work at hand builds upon these two works, improving and extending them to open it for UAV research. 

There are many more synthetic, simulated environments. Please see \cite{nikolenko2019synthetic} for an overview.

\subsubsection{Data Sets Taken on UAVs}

The first large-scale real-world object detection data sets taken on UAVs were VisDrone \cite{fan2020visdrone} and UAVDT \cite{du2018unmanned}. Other data sets for object detection and tracking emerged with many different foci \cite{pei2019human,mueller2016benchmark,hsieh2017drone,mundhenk2016large,li2017visual,krajewski2018highd,van2014nature,ofli2016combining,varga2021seadronessee,bozcan2020air}.

The need for data sets caused many researchers to focus on synthetic data. Mid-Air \cite{Fonder2019MidAir} presents a synthetic data set for unstructured environments captured with the Unreal Engine \cite{qiu2016unrealcv} in combination with AirSim \cite{shah2018airsim}. They laboriously built a world with landscape and streets for drone navigation. However, their world is lifeless and does not feature any objects of interest. The Synthinel-1 \cite{kong2020synthinel} data set features synthetic data showing building footprints with segmentation masks.

An overview comparing the most important data sets is shown in Table \ref{table:comparison_datasets}.

\subsubsection{Narrowing the Sim-to-Real Domain Gap}

On the model level, many approaches apply synthetic-to-real domain adaptation \cite{hoffman2018cycada,french2017self,li2018semantic,zou2018unsupervised,chen2018domain,prakash2019structured,tobin2017domain,tsai2018learning}. By disentangling the features, these techniques aim to learn domain invariant features that lead to better transfer capabilities. A subset of these methods performs image stylization to make the synthetic images look more similar to their real-world counterparts \cite{richter2021enhancing,zhu2017unpaired}, referred to as narrowing the appearance gap.
Another domain gap arises from differences in content, called content gap \cite{kar2019meta}. It depicts the layout and types of objects featured in the synthetic world instead of in the real world.

Orthogonal to these gaps, we focus on another gap called meta gap. Meta gap depicts the imaging conditions at the time of capture, such as altitude, viewing angle and time. It can be seen as a useful abstraction of the content gap. These metadata are freely available in the synthetic engine and on an actual UAV. We leverage these metadata to align the distributions of the synthetic and real-world data set, leading to better performance and data efficiency.




\begin{table*}
\caption{Comparison with the most prominent annotated aerial object detection data sets in three domains.}
\begin{center}
	
	\begin{tabular}{c|c|c|c|c|c|c|c|c}
		Data Set  & Domain & Data Type & \# Images & Platform  & Image Widths & Altitude &   Angle &   Other meta\\
		\hline
		\hline
	
		VisDrone \cite{zhu2018visdrone} & traffic & real & 10,209 &  UAV & 960-2,000 & \ding{53} & \ding{53} & \ding{53} \\
		
		AU-AIR \cite{bozcan2020air}  & traffic & real & 32,823  & UAV & 1,920 & \checkmark & \checkmark & \checkmark  \\
		
		PreSIL \cite{hurl2019precise} & traffic & synthetic & 40,000 & car & 1,920 & -- & --& \ding{53}\\

		\bf DGTA-VisDrone & traffic & synthetic & 50,000 & UAV & 3840 & \checkmark & \checkmark & \checkmark\\

		\hline
		\hline
		
		Airbus Ship \cite{airbus-ship}  & maritime & real & 40,000 & satellite & 768 & -- & -- & \ding{53} \\

		SeaDronesSee \cite{varga2021seadronessee} & maritime & real &  5,630 & UAV  & 3,840-5,456 & \checkmark & \checkmark  & \checkmark \\	
		
		\bf DGTA-SeaDronesSee & maritime & synthetic & 100,000 & UAV & 3,840 & \checkmark & \checkmark & \checkmark\\
		
		\hline
		\hline
		
		Cattle \cite{shao2020cattle}& agriculture & real & 670 & UAV & 4,000 & \ding{53} & \ding{53} & \ding{53}\\
		\bf DGTA-Cattle & agriculture & synthetic & 50,000 & UAV & 3,840 & \checkmark & \checkmark &\checkmark\\


	\vspace{-8mm}	
	\end{tabular}
\end{center}

\label{table:comparison_datasets}
\end{table*}

\section{Experimental Setup}

\subsection{DeepGTA-UAV}

\subsubsection{Tool Description and Improvements}

\label{sec:ToolDescriptionAndImprovements}

The DeepGTAV framework as used in this work builds upon the DeepGTAV-PreSIL framework  
\cite{hurl2019precise,deepgtav_presil}, built upon the DeepGTAV framework \cite{deepgtav}.
Originally, DeepGTAV was built as a reinforcement learning environment for self-driving cars, providing functionality to interact with GTAV through a TCP-server and building upon the functionality of SkriptHookV \cite{scripthookv}.
The Python library VPilot \cite{vpilot}
interacts with DeepGTAV, which runs in GTAV. 

DeepGTA-PreSIL integrates DeepGTAV and GTAVisionExport \cite{gtavisionexport},
the technique presented in \cite{johnson2016driving}, to extract depth and stencil buffers from the rendering pipeline. With those and the world coordinates of objects extracted in GTAV, pixel-wise object segmentation data can be extracted. From those, object bounding boxes can be calculated. The DeepGTAV framework also allows extracting voxel-wise LiDAR segmentation data, which was not used in this work. The most notable improvements made in this work include:
\begin{enumerate}
	\item Increasing the ease of use and the multitude of scenarios of which synthetic data can be generated by improving the VPilot interface of
	DeepGTAV. In particular, it is now possible to freely modify the in-game position, camera position and rotation, manipulate the environment
	(time of day, weather), spawn objects (e.g. pedestrians, cars) and specify their animations. 
	\item Improving the speed and reliability of the DeepGTAV framework (to obtain almost no overhead, compared to running GTAV natively).
	\item Allowing the extraction of metadata of the generated data (like time of day, height, camera angle, ...).
	\item Providing multiple easily comprehensible and modifiable data generation scripts for several airborne scenarios and different strategies of metadata distribution.
	\item Modifications to capture 4k image data (although we did not analyze the influence of high resolution data). 
\end{enumerate}

In this work, those adaptations were used to capture object detection data from a UAV perspective (in comparison to an autonomous car scenario in
previous works \cite{johnson2016driving, hurl2019precise}).

\subsubsection{Modifications of GTAV}

For optimal use as a simulation environment, some modifications were made to the GTAV game files by installing the modifications \textit{Simple Increase Traffic (and Pedestrian)} \cite{increasetraffic}, \textit{No chromatic aberration lens distortion} \cite{nochromatic} and \textit{Heap Limit Adjuster} \cite{heaplimitadjuster}. 
Additionally, the automatic spawning of objects was modified to match the class distribution in VisDrone. 

Furthermore, the bounding box quality was improved by setting the game resolution to 7680x4320DSR in NVIDIA GeForce Experience \cite{geforceexperience}. This results in an upscaling of the rendering buffers to 4k, which is needed to obtain pixel-perfect object segmentation data for 4k images.

\subsection{Data Set Generation}
To assess the usefulness of synthetic data as training data for real-world scenarios, particularly to assess the usefulness of DeepGTAV and its adaptability in this context, we examined three different object detection scenarios.

In choosing those three scenarios, we are bound to existing real-world data sets to assess the real-world performance. We find that the data sets VisDrone \cite{zhu2018vision}, SeaDronesSee \cite{varga2021seadronessee} and Cattle \cite{shao2020cattle} are very popular in their respective application domain and therefore choose these. VisDrone aims for traffic surveillance in Asian cities with very crowded scenes. SeaDronesSee's application domain is search and rescue in open water featuring swimmers and boats, with the main challenges being reflective regions, shadows and waves or seafoam. Finally, Cattle aims to bring autonomous vision systems to agriculture (cattle detection).

Using the DeepGTAV framework, we generate synthetic training data for these scenarios by specifying VPilot data generation scripts for each scenario. In the following, we briefly describe the different capturing procedures employed in the VPilot scripts used to generate the synthetic data sets.

For the generation of all synthetic data sets, the game world of GTAV is systematically
traversed. New images are captured with one frame per second to obtain mainly distinct images. We export the 4k images but discard the segmentation and depth maps to focus on pure object detection. Along with every frame, we export the corresponding ground truth bounding boxes and the meta labels, i.e. altitude, principal axes (yaw, pitch, roll of camera rotation), time of the day and weather state. 

For the random traversals of the game world, the camera height and angles were varied.
Additional in-game objects were spawned (e.g. vehicles, pedestrians). See Table \ref{tab:datagenerationsettings} for details. For example, in Cattle, every two seconds, four cows are spawned $50-250$m in front of the camera with a left-right offset of $-160-160$m.
The objects were spawned in the in-game traffic and pathfinding and were despawned after 200 seconds.
Additionally, a new random travel location in this area was chosen every 60 seconds to prevent the in-game navigation from getting stuck.

Concise descriptions of the data sets are given in Tables \ref{table:comparison_datasets} and \ref{table:datasetsizes}.

\begin{table}[]
    \centering
    \caption{Data set generation settings.}
    \begin{tabular}{cccc}
        \toprule
        DGTA- &  Cattle &  SeaDronesSee & VisDrone \\
        \midrule
        \begin{tabular}{c}
              GPS  \\
             $\times 10^3$ 
        \end{tabular}  & 
        \begin{tabular}{c}
             $[0, 17]\times $  \\
             $[13, 22]$ 
        \end{tabular}
        & \begin{tabular}{c}
             $[-28, -18]\times $  \\
             $[-25, -13]$ 
        \end{tabular} & 
        \begin{tabular}{c}
             $[-12, 14]\times $  \\
             $[-22, 13]$ 
        \end{tabular}
        
        \\
        \hline
        
        Altitude & $10$-$80$m & $0$-$80$m & $0$-$40$m\\
        \hline
        Cam Pitch & $20$-$90$° & $20$-$90$° & $20$-$90$°\\
         \hline
        \begin{tabular}{c}
             Manual \\
             spawn 
        \end{tabular}
        & 
         \begin{tabular}{c}
             4$\times$cow@2s   
        \end{tabular}
        & 
        \begin{tabular}{c}
             4$\times$ppl@2s\\
             4$\times$boat@2
        \end{tabular}
         & 
         \begin{tabular}{c}
             3$\times$bike@8 \\
             1$\times$motor@8
        \end{tabular}
        \\
        \hline
        \begin{tabular}{c}
             Spawn y \\
        \end{tabular} & $50$-$250$m & $50$-$250$m & $50$-$150$m\\
         \hline
         \begin{tabular}{c}
        
             Spawn x \\
        \end{tabular} & $-160$-$160$m &$-160$-$160$m & $50$-$150$m\\
        \bottomrule
    \end{tabular}
    \vspace{-5mm}
    \label{tab:datagenerationsettings}
\end{table}

\subsection{Models and Training setup}

As object detection models, we take two one-stage real-time detectors, EfficientDet-$D0$ (E.-$D0$) \cite{tan2020efficientdet} and Yolov5 \cite{redmon2016you, redmon2017yolo9000, redmon2018yolov3, bochkovskiy2020yolov4, glennjocher2020}, both being on the forefront of real-time object detectors as measured by their performances on the COCO test set \cite{lin2014microsoft,paperswith-code}. In particular, EfficientDet-$D0$ is the state-of-the-art model for real-time detectors on large-scale UAVDT \cite{du2018unmanned} traffic surveillance data set \cite{kiefer2021leveraging}. For that, we use the implementation from \cite{EfficientDetSignatrix} with an image size of 2176px width and anchor scales of (0.3 0.5 0.7).

Yolov5 \cite{glennjocher2020} is a state of the art implementation of the Yolo object detection model implemented with multiple improvements to the Yolo framework that have been found in recent years. In this work, we used the unmodified YOLOv5m6 implementation of Yolov5 in release v5.0 \cite{glenn_jocher_2021_4679653} with an image size of 1280x1280px and a batchsize of 48. Unless otherwise specified, we used the provided weights pre-trained on COCO \cite{lin2014microsoft}. 

Furthermore, as a two-stage detector we take the best performing single-model (no ensemble) on VisDrone from the workshop report \cite{zhu2018visdrone} (DE-FPN), i.e. a Faster R-CNN (F.R.) with a ResNeXt-101 64-4d \cite{xie2017aggregated} backbone (removing P6), which is trained using color jitter and random image cropping. The anchor sizes and strides are decreased to (16, 32, 64, 128, 256) and (4, 8, 16, 32, 64).

We measure the models performances on the popular mean average precision metric with overlap 0.5, i.e. mAP@0.5 \cite{lin2014microsoft}. As we are interested in real-world performance, we test on the test set of the real-world data set unless indicated otherwise.

\section{Experimental Evaluation}


First, we conducted different experiments to show that synthetic training data could yield good real-world performance or improve the performance of a real-world object detector. Then we conducted further ablation studies to examine different factors that could influence or modulate the positive effect of synthetic training data. 

On a general level, we wanted to obtain actionable advice for an engineer using synthetic data to train an object detector, which factors should be examined with emphasis and which factors could be ignored. Such factors could be the graphics quality of the simulation environment or the alignment of synthetic and real height distributions. 

In the following, those experimental conditions and their results will be described. For better clarity, we discuss the design of each ablation study and its result individually.

 \begin{table}
	\centering
	\caption{Performance of object detectors for different training strategies given by mAP@50. }
	\begin{tabular}{c|c|c|c|c}

		& \multicolumn{1}{c|}{Data set} & \multicolumn{1}{c|}{Synthetic} & \multicolumn{1}{c|}{Real} & \multicolumn{1}{c}{SyntheticToReal} \\
		\hline
		\hline
		\parbox[t]{2mm}{\multirow{3}{*}{\rotatebox[origin=c]{90}{E.-$D0$}}} & Cattle & 29.2 & 78.4 & \bf 85.8 \\
		& SeaDronesSee & 10.3 & 36.3 & \bf  38.8\\
		& VisDrone & 1.2 & 24.6 & \bf 27.2\\
		\hline
		\parbox[t]{2mm}{\multirow{3}{*}{\rotatebox[origin=c]{90}{F.R.}}} & Cattle & 38.8 & 90.5 & \bf 91.5\\
		& SeaDronesSee & 14.6 & 54.7 &\bf 59.0\\
		& VisDrone & 2.4 & 48.6 & \bf 51.2\\
		\hline
		\parbox[t]{2mm}{\multirow{3}{*}{\rotatebox[origin=c]{90}{YOLO}}} & Cattle &  64.2 & \bf 88.8 & 86.9\\
		& SeaDronesSee & 10.5 & 55.8 & \bf 60.3\\
		& VisDrone & 10.2 & 43.9 & \bf 45.0\\
	\end{tabular}
	\label{table:performance_overall}
\end{table}

\subsection{General Benefit of Synthetic Data in UAV Object Detection}

The main goal of this work is to examine the usefulness of synthetic training data to train object detectors from scratch or improve the performance of object detectors trained with real-world data.

From this goal, there naturally arise three conditions which we want to compare. First, as a baseline, we observe the performance of the object detector on the real-world data set. Second, we observe the performance of the object detector trained only on an entirely synthetically generated data set. Finally, we observe the performance of an object detector which is first pre-trained on a synthetic data set and then transfer-trained on the real-world data set. From preliminary experiments, we found these strategies to be superior to alternative strategies, such as combined training, similar to previous literature \cite{varol2017learning}.

For all three application scenarios, the corresponding real-world training set was split into a training, validation and test set. For fully synthetic training and synthetic pre-training the data set was split into a training and a validation set, as no testing is conducted on the synthetic data. See the sizes of the different complete sets in Table \ref{fig:datasetSizes}.

\begin{table}
	\caption{\label{fig:datasetSizes} Number of images and classes. Note that (awning-)tricycle is abbreviated as (a.-)tri., life jacket as LF and people as ppl.}
	\begin{tabular}{c|c|c|c|c}
	    & \multicolumn{3}{c}{Number of Images} & \\
		Data Set  & Train & Val & Test & Classes \\
		\hline
		\hline
		VisDrone & 6471 & 548 & 1610 & 
		\begin{tabular}{c}ppl.,bike,car,truck,van \\  
		motor,tri.,a.-tri.,bus
		\end{tabular} \\
		\hline
		\begin{tabular}{c}
		     Sea-  \\
		      DronesSee
		\end{tabular} & 2975 & 859 & 1796 & 
		\begin{tabular}{c}swimmer,floater,boat \\  
		swimmer$^\dagger$,floater$^\dagger$,LJ
		\end{tabular}
		\\
		\hline
		Cattle & 402 & 134 & 134 & Cow \\ 
		\hline
		\hline
		 
		\begin{tabular}{c}DGTA- \\  
		VisDrone
		\end{tabular}
		& 40000 & 10000 & -- & 
		\begin{tabular}{c}ppl.,bike,truck \\  
		car,motor,bus,van
		\end{tabular}\\
		\hline
		\begin{tabular}{c}DGTA-Sea- \\  
		DronesSee
		\end{tabular} & 90000 & 10000 & -- & \begin{tabular}{c}swimmer,floater,boat \\  
		swimmer$^\dagger$,floater$^\dagger$
		\end{tabular}\\
		\hline
		\begin{tabular}{c}DGTA- \\  
		Cattle
		\end{tabular} & 40000 & 10000 & --  & Cow \\
		\hline
	\end{tabular}
	\label{table:datasetsizes}
\end{table}

\subsubsection*{Findings}


Table \ref{table:performance_overall} shows that purely training on synthetic data can already provide minimal working solutions. While all object detectors perform well on the simple data set Cattle (29.2-64.2 mAP@50), the accuracies on VisDrone and SeaDronesSee are far lower. This is partly due to missing classes in the corresponding synthetic data sets ((awning-)tricycle in VisDrone and life jacket in SeaDronesSee). 
However, another apparent factor is the difference in the appearance of certain classes from the synthetic to the real data set. For example, see Figure \ref{fig:intro_img} to compare the same classes in the synthetic and real data set. The default appearances of classes vary in some cases significantly. This appearance gap may be narrowed by manually editing the appearance of classes to resemble real-world objects. 
Despite these challenges, synthetic pre-training with subsequent transfer training boosts performance on VisDrone and SeaDronesSee significantly across all models. On the SeaDronesSee evaluation benchmark \cite{seadronesseebenchmark}, we can even achieve state-of-the-art performance by surpassing the best model by +5.6 mAP@50.  While the performance improvement is not as apparent on Cattle, pre-training on synthetic data helps for EfficientDet-$D0$ and Faster R-CNN.

These experiments show that although synthetic data sets can not replace corresponding real-world data sets, they enhance the detection performance. Even the models from \cite{varga2021seadronessee} can be beaten just by synthetic pre-training.

\subsection{Effect of Data Set Sizes}
\label{sec:EffectDatasetSize}

We wanted to test the influence that the data set sizes had on the performance in this context. The intuitive hypothesis was that we would observe a curve of diminishing returns for larger data sets, which is typical in machine learning. We hypothesized that we would observe such diminishing returns for the size of the real-world data set, as well as for the size of the synthetic (pre-training) data set. Furthermore, we hypothesized that the effect of synthetic pre-training would be more emphasized when using a smaller real-world data set.

To conduct this ablation, we varied the size of the real-world data set and of the synthetic data set for VisDrone and SeaDronesSee training.

\begin{figure}
	\includegraphics[width=\linewidth]{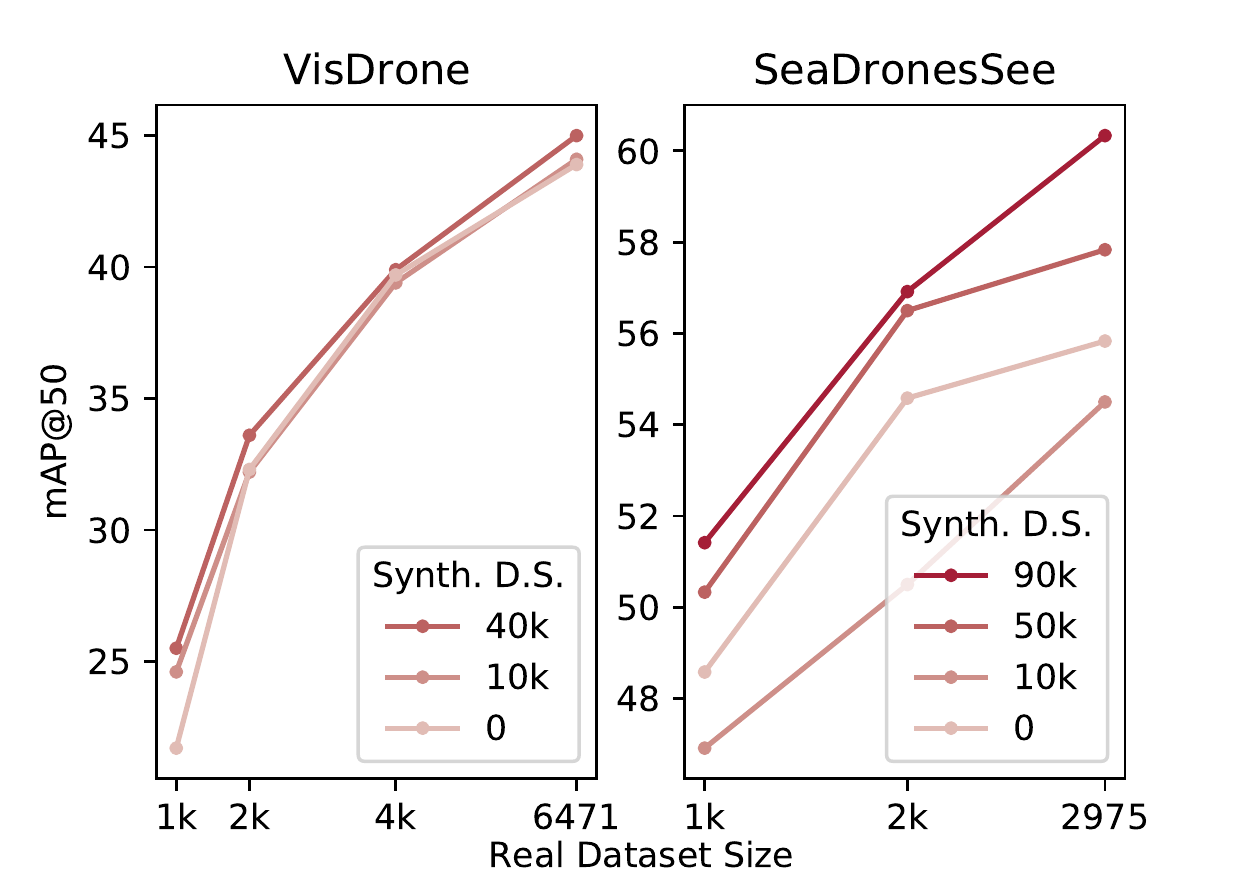}
	\caption{The effect of the real and synthetic data set size on the model performance for VisDrone and SeaDronesSee. The color shading specifies the size of the synthetic data set that was used for pre-training.}
	
	\label{fig:DatasetSizeEffect}
\end{figure}

    

\subsubsection*{Findings}

Figure \ref{fig:DatasetSizeEffect} shows the effect of different data set sizes in synthetic pre-training. As hypothesized, we observe an improved mAP@50 with an increase of the real-world data set size, and an increase of the synthetic pre-training data set size with a diminishing return for larger data sets. Furthermore, we observe the hypothesized interactive effect, such that the improvement from using synthetic pre-training data is disproportionally larger for smaller real-world data sets.

Our found effects are consistent with the numbers of those reported in \cite{hurl2019precise}, that is, improvements of about +1.0 to +5.0 mAP@50 by using additional synthetic training data. We note that in \cite{hurl2019precise}, a 3D LiDAR object detection task was examined instead of a 2D object detection task.
Compared to \cite{hurl2019precise}, in our work, those improvements are not only on one class but over the whole 10 and 6 classes of VisDrone and SeaDronesSee, respectively.

\subsection{Effect of using pre-trained weights}
\label{sec:EffectPretrainedWeights}
The use of initial weights pre-trained on large scale image data sets like ImageNet \cite{deng2009imagenet} or COCO \cite{lin2014microsoft} has become a standard for many vision machine learning tasks in recent years. We hypothesized that there could be an interaction between using such pre-trained weights and synthetic \mbox{(pre-)training}. For example, such pre-trained weights could encode invariance to image noise, which is prominent in real world images, but may be missing in synthetically generated images. The use of pre-trained initial weights would thereby allow a purely synthetically trained model to obtain the information it could not obtain from the synthetic data, thereby introducing an interactive effect. If this hypothesis were true, using pre-trained weights would improve the performance of a purely synthetically trained model disproportionally more than it improves the performance of a model trained on real world data.


\begin{table}[]
    \centering
    \caption{Effect of COCO pre-trained weights on performance given by mAP@50 evaluated on VisDrone.}
    \begin{tabular}{lrr}
        \toprule
        {} &  random initial weights &  Pre-trained on COCO \\
        \midrule
        Real                           &                  43.1 &               43.9 \\
        Synthetic                      &                  7.5 &               10.2 \\
        SyntheticToReal &                  44.7 &               45.0 \\
        \bottomrule
    \end{tabular}
    
    \label{tab:COCOPretrainingEffect}
\end{table}

\subsubsection*{Findings}

Table \ref{tab:COCOPretrainingEffect} shows the effect of using COCO pre-trained weights as initial weights for the training.
Due to the small effect sizes, we would argue that our results are inconclusive at this time. We observe a more significant absolute improvement of the mAP@50 when using COCO pre-trained weights on pure DGTA-VisDrone training than on pure VisDrone training, which would speak for the hypothesized interactive effect. However, when observing the absolute obtained mAP@50 of those different conditions, this observed difference in improvements would also be consistent with an effect of diminishing returns of improvements for a larger mAP@50. This effect would be that starting from a weaker performance baseline, the same improvement (using COCO pre-trained weights) would yield a larger improvement than when starting from a more robust baseline.

So we conclude that there is at least no strong effect where using pre-trained weights trained on real-world imagery would disproportionally improve the performance of synthetic training. However, note that there might be other factors to consider, such as training time and stability.

\subsection{Effect of Good/Bad graphics settings}
\label{sec:GraphicsQuality}

One of the parameters that one would intuitively look at when observing synthetic training data or when trying to improve the use of synthetic training data is the realism of this training data. This could be framed as closing the Sim-To-Real gap as discussed above.

A similar but not fully congruent perspective on this coming from game development is trying to improve the realism of games by improving their graphics quality. In many cases, those improvements of graphics quality come with high computational demands (higher polygon models, higher resolution textures, demanding physics simulations like particle effects and lighting) and high amounts of labor, e.g. from graphics artists. 

Therefore, from an engineering perspective, this raises the question, how well spent this effort is to obtain the final goal of increasing the synthetically trained model's performance. 

To answer this question, we conducted experiments comparing the effects of synthetic training data obtained from GTAV either set to the highest or lowest possible graphics settings. 



\begin{table}[]
    \centering
    \caption{Performance between low quality and high quality in-game settings (given by mAP@50). We compare the performance for only Real training on VisDrone against the performance with synthetic data as DGTA-VisDrone.}
    \begin{tabular}{llll}
        \toprule
        {} & Real & Synthetic & SyntheticToReal \\
        \midrule
        Real data Baseline &                        43.9 &                       - &                                 - \\
        Low Quality        &                            - &                  8.4 &                             45.1 \\
        High Quality       &                            - &                   10.2 &                              45.0 \\
        \bottomrule
    \end{tabular}
    
    \label{tab:LQHQEffect}
\end{table}

\subsubsection*{Findings}
Table \ref{tab:LQHQEffect} shows the obtained model performance using synthetic data obtained from GTAV on the lowest and highest graphics settings, respectively. We observe that for purely synthetic training, the model trained with higher graphics setting images outperforms the one trained on lower graphics quality images. This effect, however, is small.

There is no observable effect in the synthetic pre-training condition with transfer training on the real-world data set.


\subsection{Aligning Domain Distributions}
\label{sec:aligning}
Modifying the DeepGTAV tools to extract metadata allows us to automate parts of the data generation process by aligning the metadata distributions of the real and synthetic data sets. Instead of laboriously setting the correct metadata settings in the synthetic data generation process, one could fall back to the corresponding real data set to adapt the parameters automatically. Aligning the distributions may result in higher synthetic data quality or efficiency. 

For instance, in SeaDronesSee, every image is annotated with the capture time stamp. By bootstrap sampling from the time distribution, we sample a new data set of 100k synthetic images with the correct time distribution. As before, we train a Yolov5 
model and test it on the SeaDronesSee test set (Synthetic Only), and we transfer train it on SeaDronesSee (Synthetic-To-Real). 

\subsubsection*{Findings}
Table \ref{tab:MetaAlignExperiments} shows that aligning the time helps in synthetic only training and synthetic pre-training by increasing the performance over the unaligned baseline by +4.1 and +0.2 mAP@50, respectively. The performance increase is mainly due to SeaDronesSee only featuring day-time images, such that sampling synthetic night images deteriorates the performance.

Similarly, we can also automatically adjust the camera angle. The images in the Cattle data set have been taken from a downward-facing camera where a gimbal corrects for UAV angular movement. We sample from DGTA-Cattle only these images that fall into the range of these angles (with an error threshold of at most 20 degrees). This reduces the original 40k images to only roughly 10\% of the images (3,954). We train an EfficientDet-$D0$ on this subset. 
\subsubsection*{Findings}
Interestingly, the performance of the synthetic only training improves over the more extensive unaligned DGTA-Cattle training. This is likely due to the limited capacity of an EfficientDet-$D0$ model resulting in the model distributing its performance across many other angular viewpoints, which are unnecessary for the performance in this use-case. This experiment illustrates that less but more targeted data may be sufficient to reach the same performance while reducing the need to filter the data by only aligning the metadata distributions manually.



\begin{table}[]
    \centering
    \caption{Effect of time and angle alignment on performance (given as mAP@50). }
    \begin{tabular}{lrr}
        \toprule
        SeaDronesSee &  Synthetic &  SyntheticToReal    \\
        \midrule
        Only real training            &                      - &                                                    55.8 \\
        Unaligned Synthetic Pre-training &                   10.5 &                                                    60.3 \\
        Time Aligned                              &                 \bf  14.6 &                                                 \bf   60.5 \\

        \bottomrule
        \toprule 
        Cattle & & \\
        \midrule
        Only real training             &                      - &                                                    78.4 \\
        Unaligned Synthetic Pre-training &                   29.2 &                                                   \bf   85.8 \\
        Angle Aligned (Subset)                              &                 \bf  36.6 &                                                 \bf   85.8 \\

        \bottomrule
        
    \end{tabular}
    
    \label{tab:MetaAlignExperiments}
\end{table}

\section{Limitations and Conclusions}

We demonstrated that synthetic data can be leveraged for object detection on UAVs. We can improve the performance over only real training by synthetic pre-training on multiple application scenarios. Synthetic-only training yields satisfactory results, but performances are not yet competitive to real training. 

From our ablations, we conclude that the use of more synthetic training data improves the performance. The use of weights pre-trained on large scale image data sets constantly improves the performance, although we find no interactive effect with synthetic training. The graphics quality of the simulation engine appears to be important for purely synthetic training but not for synthetic pre-training. In general, metadata alignment is vital for the usefulness and data efficiency of synthetic training data. 

We hope that the adaptation of the DeepGTAV tools helps cast light on object detection on UAVs via synthetically generated footage.
In future works, the capabilities of the DeepGTAV framework to produce object segmentation data, LiDAR data and video could be leveraged.

\newpage
\bibliographystyle{IEEEtran}
\bibliography{IEEEabrv,bare_conf}

\end{document}